\documentclass[preprint,12pt]{elsarticle}

\usepackage{lmodern}
\usepackage[T1]{fontenc}

\usepackage{amssymb}
\usepackage{amsmath}

\usepackage{subfig}
\usepackage{bm}
\usepackage{xspace}
\usepackage{ctable}
\usepackage{longtable}
\usepackage{newtxmath}
\usepackage{url}
\usepackage{algorithm}
\usepackage{algpseudocode}
\usepackage{threeparttable}
\usepackage{tabularx}

\makeatletter
\DeclareRobustCommand{\ie}{\textit{i.e}\@ifnextchar.{}{.\xspace}}
\DeclareRobustCommand{\eg}{\textit{e.g}\@ifnextchar.{}{.\xspace}}
\DeclareRobustCommand{\af}{\textit{ac.f}\@ifnextchar.{}{.\xspace}}
\DeclareRobustCommand{\afs}{\textit{ac.f}s\@ifnextchar.{}{.\xspace}}

\newcommand{\oned}{\textit{1d}\xspace}

\newcommand{\threed}{\textit{3d}\xspace}

\newcommand{\adhoc}{\textit{ad hoc}\xspace}

\makeatother

\journal{Information Sciences}

\begin{document}

\begin{frontmatter}

\title{Robust Deep Network Learning of Nonlinear Regression Tasks by Parametric Leaky Exponential Linear Units (LELUs) and a Diffusion Metric}

\author{Enda D.V. Bigarella}
\ead{enda.bigarella@gmail.com}
\affiliation{organization={Independent Researcher},
            addressline={Starnberger Weg 7}, 
            city={Gilching},
            postcode={82205}, 
            country={Germany}
            }

\begin{abstract}%
This document proposes a parametric activation function (\af) aimed at improving multidimensional nonlinear data regression.
It is a established knowledge that nonlinear \afs are required for learning nonlinear datasets.
This work shows that smoothness and gradient properties of the \af further impact the performance of large neural networks in terms of overfitting and sensitivity to model parameters.
Smooth but vanishing-gradient \afs such as ELU or SiLU (Swish) have limited performance and non-smooth \afs such as RELU and Leaky-RELU further impart discontinuity in the trained model.
Improved performance is demonstrated with a smooth ``Leaky Exponential Linear Unit'', with non-zero gradient that can be trained.
A novel diffusion-loss metric is also proposed to gauge the performance of the trained models in terms of overfitting.
\end{abstract}


\end{frontmatter}

\section{Introduction}
\subsection{Perspectives and Motivation}
It is typical in engineering applications that a modelled system is multidimensional, \ie, a representative output parameter of the system depends on many system operational variables.
The model might also be structured following a mesh-like combination of the system variables, which were input on a computational model or a test bench for the data generation.
In the example to be addressed in this document, a metric of an electric motor efficiency is dependent on three variables, representative of the motor speed, power, and electric feed.
This seemingly simple problem has proven to be surprisingly difficult to be trained for data regression.
In an effort with Gaussian Process Regression (GPR) \cite{bigarellaPreventionOverfittingMeshStructured2025}, the author developed a loss based on a diffusion operator to detect and avoid overfitting resulting from the highly nonlinear training landscape.

An argument in \cite{mingardDeepNeuralNetworks2025} stating that deep-learning methods naturally provides a Occam's Razor behaviour has instigated the author to apply a neural network (NN) to the problem.
There was an expectation that this property would naturally avoid overfitting as seen with the GPR models.
The nature of the motor model has also raised some learnings in this deep learning context and are herein reported.

The scientific and engineering communities are interested in modelling techniques that are robust to parameters and that provide good generalisation.
The selection of the activation function (\af) of a NN heavily impacts such objectives, and it is known that \textit{nonlinear} \afs are necessary to model nonlinear datasets \cite{goodfellowDeepLearning2016}.
Furthermore, \textit{smoothness} (\eg, the classical ELU \cite{clevertFastAccurateDeep2016}) as well as \textit{trainable/adaptive} coding properties for \afs receive consistent attention in the literature, and the reader is referred to the thorough summary in \cite{jagtapHowImportantAre2022} for further information.
More complex approaches such as \afs based on GPRs \cite{urbanGaussianProcessNeurons2017}, rational functions \cite{boulleRationalNeuralNetworks2020}, Kronecker Neural Networks \cite{jagtapDeepKroneckerNeural2022}, or even on model structural changes \cite{apicellaSimpleEfficientArchitecture2019}, have also been proposed.
Of importance to the current work, several trainable variants are available solely for ReLU and ELU (again summarised in \cite{jagtapHowImportantAre2022}).

The choice of an \af can in fact be overwhelming and other broad-scope reviews are available in \cite{apicellaSurveyModernTrainable2021, szandalaReviewComparisonCommonly2021}.
However, the literature generally focuses on large models for classification problems whereas regression problems are typically small.
Even though such reviews provide interesting background for further exploration, regression tasks can be more demanding than classification since they require continuity.
An interesting example of such difficulties is a compromise termed ``benign overfitting'' \cite{haasMindSpikesBenign2024}, in which localised spikes near noisy training points are accepted for a model that generalises well elsewhere.

In the current work, a dense feedforward NN is developed to learn the aforementioned motor model.
Relatively large depths and widths are required to properly learn such nonlinear problem \cite{schmidt-hieberNonparametricRegressionUsing2020, Caruana2001Overfitting}.
As shown in Sect.\ \ref{sheselectric}, over-expression between adjacent data points is observed with classical \afs and the use of regularisation to avoid such patterns incur in larger training errors.
In special, the regions where two of the variables are fast changing but the third one is not, seemed to exacerbate the issue.

A motivation to pursue a deeper understanding of the observed issues comes from this remark in \cite{dugasIncorporatingSecondOrderFunctional2000}, that ``a priori knowledge of a particular task into a learning algorithm (...) generally improves performance, if the incorporated knowledge is relevant to the task.''
The author proposes an \adhoc metric of the flexibility of the \af as a ratio of its maximum and minimum gradients.
In considering this metric, it can be observed that highly flexible \afs are prone to over-expressing in large models, whereas the conventional linear unit has zero flexibility and thus only represents linear functions (a common knowledge \cite{goodfellowDeepLearning2016}).

\subsection{Contributions of the Work}
The current work proposes a parametric, trainable flexibility that enhances the a priori ``fitness'' of an \af for more generalisable nonlinear learnings.
This outcome is herein achieved with a simple \af definition, termed Leaky Exponential Linear Unit (Leaky ELU, or LELU).
Importantly, the focus here is broader in the sense that it focuses on a set of behavioural features of an \af for generalisation, one of them being the discussed flexibility, but also robustness (\ie, reduced dependency) with model size.
Quoting \cite{jagtapHowImportantAre2022} ``there is no rule of thumb of choosing an optimal activation function.''
Essentially, we want to decrease the effort in finding an optimal \af by providing an adaptive framework that is able to code the degree of presence of particular phenomena in the input in a smooth and ever proportional way.

Another proposition in the current work is the use of a diffusion-like metric to gauge the performance of the NN, especially in terms of overfitting.
As developed in \cite{bigarellaPreventionOverfittingMeshStructured2025}, the ``entropy-in-features'' of the training data is computed via a modified Laplace operator and considered as a true label.
Testing performed on a staggered mesh measures the entropy content of a trained model with respect to the true label, essentially meaning that a trained model should limit the amount of extra features it determines, and thus less ``noise'' stemming from overfitting.

\subsection{Organisation}
The rationale for the \af definitions are initially discussed.
The NN setup and diffusion loss definitions are then presented.
The localised overfitting behaviour is demonstrated on canonical unidimensional problems, with model sizes representative of that for the motor model.
It is shown that classical \afs have mixed performance in the problems whereas the proposed LELU is robust to model size.
The observed properties of the LELU are further demonstrated in the motor model.

\section{Activation Functions} \label{afs}
\begin{figure}[t!]
	\centering
	\includegraphics[clip, width=10cm]{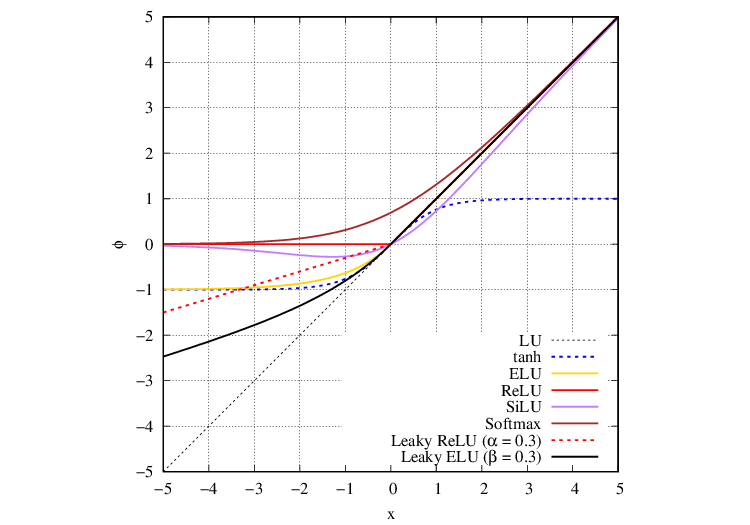}
	\caption{Graph of several activation functions.}
	\label{activ_fks}
\end{figure}

The author herein proposes an \adhoc score of the ``flexibility'' (term inspired by \cite{hawkinsProblemOverfitting2004}) of an \af to serve as a proxy of its behaviour in data regression tasks.
The flexibility score $\eta(\phi)$ is proportional to the ratio of the maximum and minimum slopes of an \af $\phi$, such that
\begin{equation} \label{flexaf}
    \eta(\phi) = 1 - \frac{\min(\phi')}{\max(\phi')}.
\end{equation}
This scores is zero for inflexible \afs, such as the Linear Unit (LU), which can only represent linear functions.
On the other extreme, an \af with score 1 is highly flexible.
Table \ref{flexaftable} shows the score for the \afs plotted in Fig.\ \ref{activ_fks}, and includes a wide range of original to more modern \afs.

\begin{table}[h!]
	\centering
	\begin{threeparttable}
	\caption{Flexibility score of several activation functions by application of Eq.\ \ref{flexaf}. The parametrizable activation functions can affect their flexibility score (examples in Fig.\ \ref{activ_fks}).} \label{flexaftable}
	\begin{tabularx}{0.8\textwidth}{>{\hsize=.75\hsize}X|>{\hsize=.25\hsize}X}
		\toprule
		\textbf{Activation} $\mathbf{\phi}$ & $\eta(\phi)$ \\
		\midrule
		LU         	& 0 \\
		tanh        	& 1 \\
		ReLU        	& 1 \\
		ELU \cite{clevertFastAccurateDeep2016}						& 1 \\
		Leaky ReLU \cite{maasRectifierNonlinearitiesImprove2013, heDelvingDeepRectifiers2015}	& $1-\alpha$ \\
		SiLU* \cite{hendrycksGaussianErrorLinear2023} (also, Swish) 	& 1.1 \\
		Softplus \cite{dugasIncorporatingSecondOrderFunctional2000}	& 1 \\
		LELU			& $1-\beta$ \\
		\bottomrule
	\end{tabularx}
	\begin{tablenotes}
	\item \small{*The term SiLU instead of Swish \cite{ramachandranSearchingActivationFunctions2017} is adopted for consistency with the Keras library. For the current purposes, the other modern option Mish \cite{misraMishSelfRegularized2020} showed very similar performance to SiLU/Swish.}
	\end{tablenotes}
	\end{threeparttable}
\end{table}

In the current work, it is shown that for \textit{nonlinear regressions tasks} the following features of an \af negatively impact the model performance:
\begin{itemize}
\item a higher flexibility score is more likely to present overfitting,
\item discontinuous functions impart discontinuity into the trained data,
\item vanishing gradient impairs training, irrespective if the function saturates at non-zero values \cite{maasRectifierNonlinearitiesImprove2013},
\item absence of negative values is too restrictive for neuron activation for smooth functions and limits their learning.
\end{itemize}
This motivates the proposal of a parametric \af with a trainable parameter that overcomes these limitations and thus allows for faster and more robust training for nonlinear regression tasks.
The family of activation functions, termed ``Leaky Exponential Linear Unit'' (Leaky ELU, or LELU), is given by
\begin{equation}
	\text{LELU}(x, \beta) = 
	\begin{cases}
		~ x & \text{if } x > 0 \\
		~ \exp((1 - \beta) x) - 1 + \beta~x & \text{if } x \leq 0
	\end{cases}
\end{equation}
where the trainable parameter $\beta$ controls the slope of the activation function in the negative region (see Fig.\ \ref{activ_fks}).
Interestingly, the elastic ELU in \cite{kimElasticExponentialLinear2020} proposes a change in the slope of the \af however on the positive, right-hand side.

The parametric LELU is constructed by imposing the following objectives for its features and behaviours
\begin{enumerate}
\item monotonically continuous,
\item non saturating (non-null gradients),
\item to push the unit mean activation closer to zero.
\end{enumerate}
These requirements are imposed onto the \af construction to enhance the performance for nonlinear regressions tasks.
The regression results show that this set of features inherently produces a behaviour similar to batch normalisation without the need for regularisation, as for instance explicitly remarked in \cite{houNeuralNetworksSmooth2016, boulleRationalNeuralNetworks2020, huMeasuringModelComplexity2020}.

\section{Model Description}
The general aspects of the deep-learning model are initially presented.
In the sequence, the diffusion-loss metric formulation is described.

\subsection{Neural Network Configuration}
\begin{figure}
	\centering
	\includegraphics[clip, width=8cm]{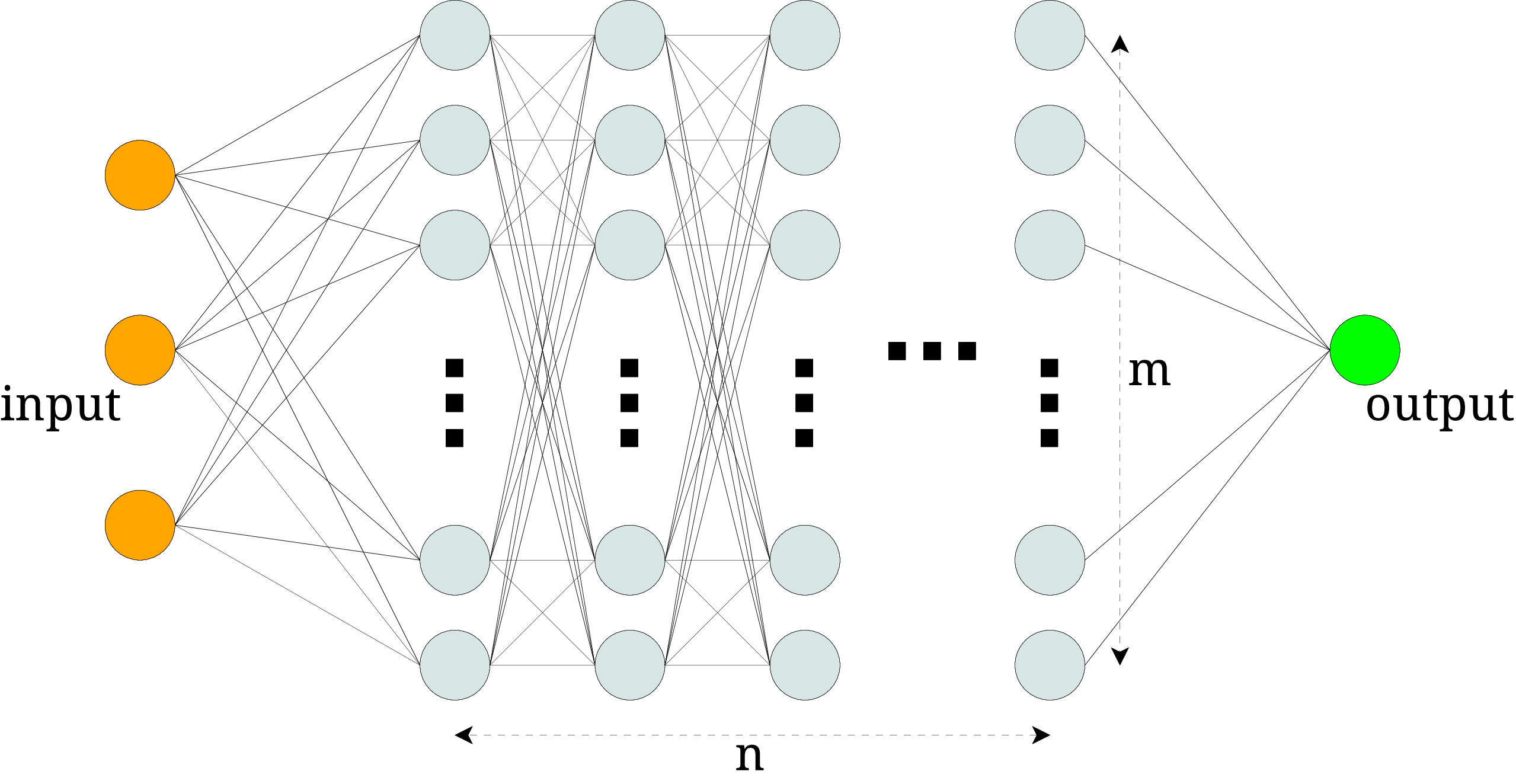}
	\caption{A vanilla feedforward neural network of depth $n$ and width $m$, with three inputs and one output. The output layer is linear.}
	\label{nnsetup}
\end{figure}

A fully-connected feedforward NN is set up with $n$ hidden layers each containing $m$ neurons, and a linear-uniform single-neuron output, as sketched in Fig.\ \ref{nnsetup}.
The NN is implemented in Python using the Keras library \cite{cholletKeras2015} and trained to minimise the mean absolute error (MAE) employing the Adam optimizer \cite{kingmaAdamMethodStochastic2017}, as summarised in Algorithm \ref{algo}.

\begin{algorithm}[H]
	\caption{Training procedure of the artificial neural network.}\label{algo}
	\begin{algorithmic}[1]
	\Require Training data $(\vec{x}, f(\vec{x}))$.
	\Comment{From system models or generated from canonical functions.}
	\Ensure Trained neural network and diffusion-loss metric.
	\Procedure{NN Setup}{}
    		\State Define the NN architecture (depth and width, see Fig.\ \ref{nnsetup}).
    		\State Configure learning rate annealing and number of epochs for the Adam optimizer.
    		\State Set activation functions.
    		\Comment{Regularisation applied when intended.}
    		\State Initialize weights with chosen initializer.
	\EndProcedure
	\While{max epoch not reached}
    		\State Train the NN by back-propagation to minimise the MAE$(f_{predicted}, f)$ at the training points.
	\EndWhile
	\Procedure{Diffusion metric}{}
	\Comment{Detailed in Sect.\ \ref{section_method}.}
    		\State Make predictions at staggered mesh nodes and at training points.
    		\State Compute diffusion terms at the training points according to Eq.\ \ref{fulldiffuterms}.
    		\State Compute the diffusion loss applying Eq.\ \ref{diffuloss}.
	\EndProcedure
	\end{algorithmic}
\end{algorithm}

The minimisation of the MAE has provided better results for the regression tasks when compared to a mean squared error (MSE).
The learning rate is step-annealed based on the MAE loss and the weights are initialised by proper techniques, with more details later provided for each training case.

Regularization in regression tasks is essentially a trade-off: it helps prevent overfitting and can improve training stability, but by definition the minimum achievable training error is limited by the strength of the regularization.
Therefore, no regularisation is used as default in the current work, nonetheless this aspect is also herein assessed since it is a common practice.

\subsection{Diffusion-Loss Metric} \label{section_method}
The use of the diffusion loss in Algorithm \ref{algo} provides an independent \textit{testing metric} of performance and also allows for \textit{training on the full dataset}, as discussed in the current subsection.
The diffusion-loss formulation herein applied follows the more complete description in \cite{bigarellaPreventionOverfittingMeshStructured2025}.
This methodology allows for the creation of a hierarchical testing metric from the training data field without requiring splitting a training dataset into testing sets.
Training is performed on the complete available dataset and offers best chance for generalisation. 

\begin{figure}
	\centering
	\includegraphics[width=8cm]{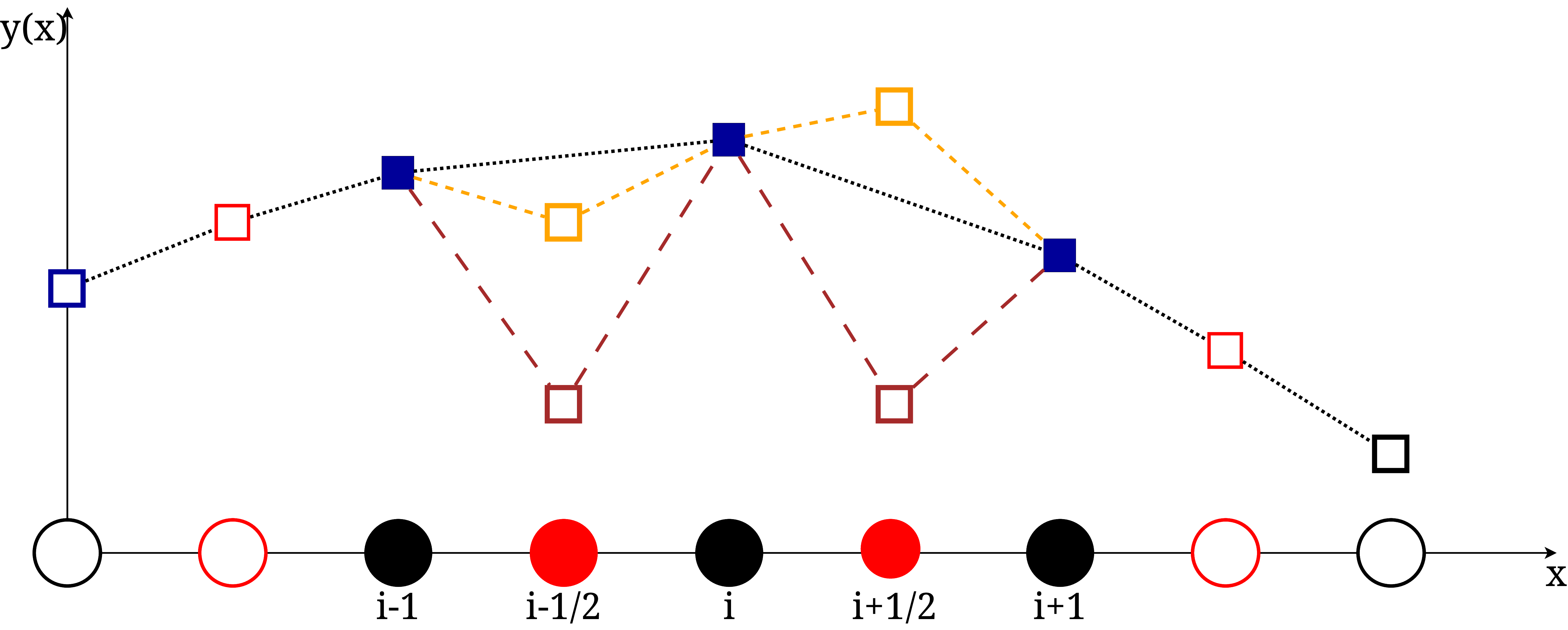}
	\caption{A one-dimensional mesh: the function values are shown by squares and the mesh nodes by circles. The original training mesh is shown as black nodes and the created staggered nodes, in red with half indices. Brown and orange squares show possible oscillations around node $i$.}
	\label{staggered1d}
\end{figure}

The method for identifying oscillations uses a measure of the diffusion of the training data represented by a modified Laplace operator $\nabla$.
This sensor in a \oned finite-difference context is computed as follows
\begin{equation} \label{truescore1d}
\nabla y_i = \frac{1}{\Delta^2} \frac{\big| y_{i+1} - 2 y_{i} + y_{i-1} \big|}{y_{i+1} + 2 y_{i} + y_{i-1}},
\end{equation}
for the $i$-th point of the mesh in Fig.\ \ref{staggered1d}, with $\Delta$ representing the mesh spacing, and $y$ a positive property.
In the currently proposed method, this diffusion sensor applied to the \textit{training} data serves as the \textit{true label}.

In order to create a suitable \textit{testing metric} sensitive to overfitting in between training points, \textit{staggered} nodes centred between the training points are created (shown in red in Fig.\ \ref{staggered1d}).
An updated diffusion sensor $\tilde\nabla$ applied to these staggered nodes, indexed by half indices, is proposed
\begin{equation} \label{staggeredlaplacian}
\tilde\nabla y_i = \frac{1}{3~(\Delta/2)^2} \frac{\big| \hat{y}_{i+1} - \hat{y}_{i+\frac{1}{2}} - \hat{y}_{i-\frac{1}{2}} + \hat{y}_{i-1} \big|}{\hat{y}_{i+1} + \hat{y}_{i+\frac{1}{2}} + \hat{y}_{i-\frac{1}{2}} + \hat{y}_{i-1}},
\end{equation}
with $\hat{y}$ representing \textit{predicted} function values, $\hat{y}_i = f(x)_i$, with $\hat{y} = y$ expected at the training points.
This stencil is able to identify the ``ringing/wiggling'' features around the central node $i$.
Therefore, deviations between a testing sensor (Eq.\ \ref{staggeredlaplacian}) and the true score (Eq.\ \ref{truescore1d}) can thus be used as testing metric in model training.

\begin{figure}
  \centering
  \includegraphics[width=0.625\textwidth]{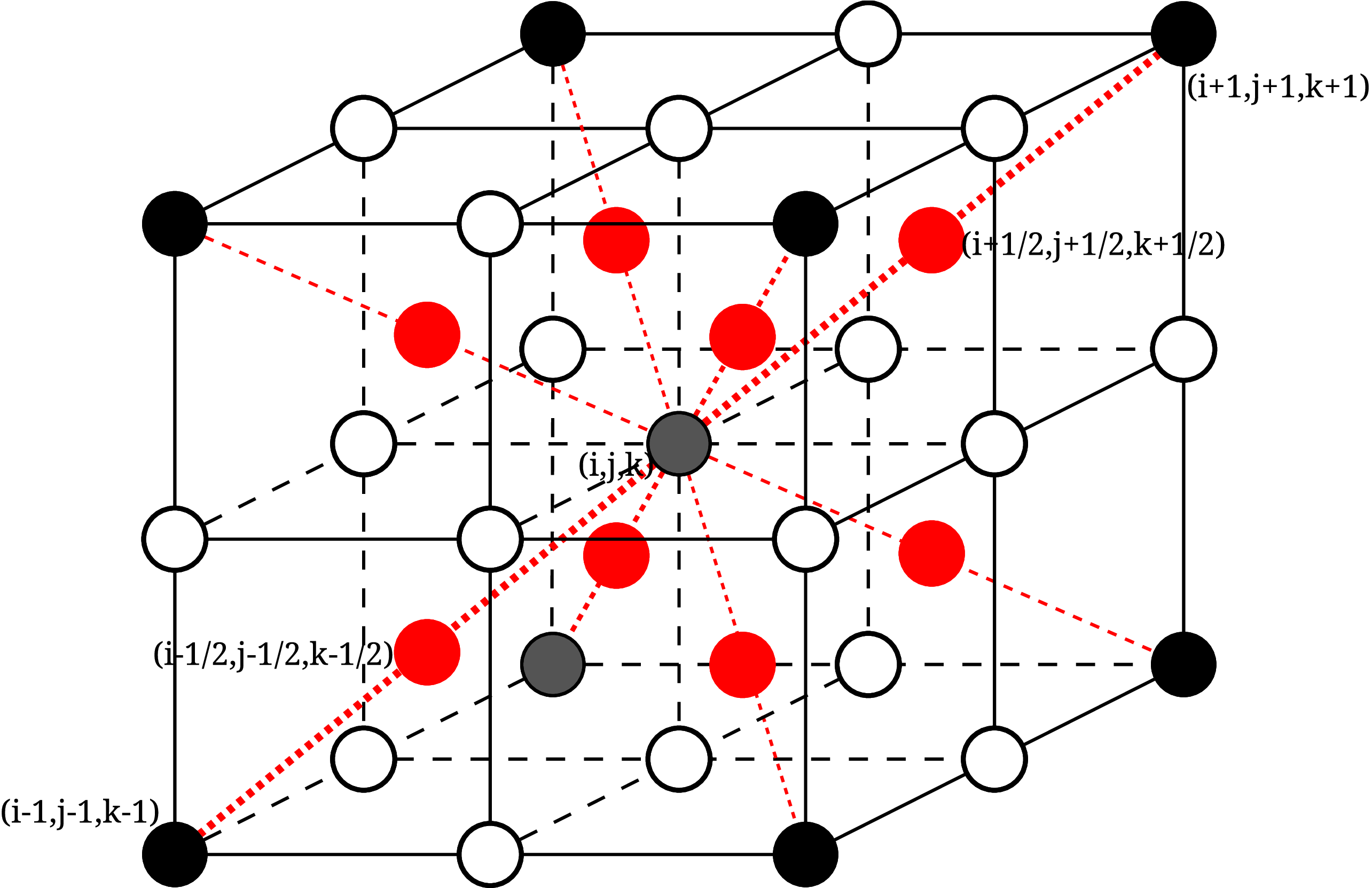}
  \caption{The original multidimensional training mesh shown as black points and the created cell-centred staggered points shown in red, with half indices. The diffusion sensors are computed along the centre-crossing diagonals shown in red. Filled symbols are used for the sensor computations for the central node ($(i,j,k)$ in \threed, for instance).}
  \label{staggereddiag}
\end{figure}

The creation of a staggered mesh in one dimension as shown in Fig.\ \ref{staggered1d} is straightforward since a central node only has a single neighbour to either its side.
On a multidimensional case, an approach based on a single staggered mesh built on centroids of the original mesh cells is proposed.
In this approach, we are required to use the centre-crossing diagonals (composed of pairs of antipodal points) as an \textit{ad-hoc} means to compute Laplace operators.
The resulting staggered mesh for a \threed case is shown in Fig.\ \ref{staggereddiag}.

The \textit{staggered} Laplace-operator diffusion sensor written for a centre-crossing diagonal $p$ of the node $(i,j,k)$ in Fig.\ \ref{staggereddiag} is given by
\begin{equation} \label{staggeredmdlaplacian}
\tilde\nabla y_{i,j,k}^{p} = \frac{1}{3~(\Delta/2)^2} \frac{\big| y_{i+1,j+1,k+1} - y_{i+\frac{1}{2},j+\frac{1}{2},k+\frac{1}{2}} - y_{i-\frac{1}{2},j-\frac{1}{2},k-\frac{1}{2}} + y_{i-1,j-1,k-1} \big|}{y_{i+1,j+1,k+1} + y_{i+\frac{1}{2},j+\frac{1}{2},k+\frac{1}{2}} + y_{i-\frac{1}{2},j-\frac{1}{2},k-\frac{1}{2}} + y_{i-1,j-1,k-1}},
\end{equation}
and for the true label
\begin{equation} \label{mdlaplacian}
\nabla y_{i,j,k}^{p} = \frac{1}{\Delta^2} \frac{\big| y_{i+1,j+1,k+1} - 2 y_{i,j,k} + y_{i-1,j-1,k-1} \big|}{y_{i+1,j+1,k+1} + 2 y_{i,j,k} + y_{i-1,j-1,k-1}}.
\end{equation}
The complete \threed diffusion sensors for the node $(i,j,k)$ are given by the summation over all centre-crossing diagonals resulting in
\begin{equation} \label{fulldiffuterms}
\nabla y_{i,j,k} = \sum_{p=1}^{4} \nabla y_{i,j,k}^{p}~, ~~~\tilde\nabla y_{i,j,k} = \sum_{p=1}^{4} \tilde\nabla y_{i,j,k}^{p}.
\end{equation}
Note that the diffusion sensors are only computed at the interior of the original mesh, where the complete stencil is available.
The loss of the diffusion sensor is finally computed as the MSE by
\begin{equation} \label{diffuloss}
\mathrm{MSE}_{diffusion} = \frac{1}{E} \sum_{e=1}^{E} (\nabla y_{e} - \tilde\nabla y_{e})^2,
\end{equation}
where $E$ is the internal nodes of the training mesh.
The reader is referred to \cite{bigarellaPreventionOverfittingMeshStructured2025} for further and more general discussions and information about the methodology as well as snippets of code implementation.

\section{Results and Discussions}
We are interested in multidimensional problems that typically require large models to capture all their features.
The author argues that using canonical problems with isolated features and fitting small NNs to them do not guarantee generalisation of the approach for those large, real-world problems.
We want to be able to code the degree of presence of series of particular phenomena that might compose large problems.
Therefore, in this work, we take large NNs that are required to represent large, complex models, and apply them to canonical problems in search of \afs that offer robustness whilst coding the degree of presence of different and composed phenomena.
In essence, the context within is models with excess capacity surpassing network size guidelines \cite{rynkiewiczGeneralBoundOverfitting2012} to better fit regions of high non-linearity \cite{Caruana2001Overfitting} and that likely exceed the sample size \cite{schmidt-hieberNonparametricRegressionUsing2020, constantinescuApproximationInterpolationDeep2024}.
The goal is to provide an \af that works well regardless of model size and that can scale effectively from small to large networks, thereby enabling potentially more robust and less time consuming regression tasks.

\subsection{Unidimensional Problems}
The proposed unidimensional problems\footnote{Source files available in \url{http://github.com/ebiga/nn_1d_overfit_ex}} represent certain important features of the target motor model.
A hyperbolic tangent function is chosen for its non-monotonic change of slope, whereas an exponential function is chosen for its large and fast change of properties.
Numerical experiments with NNs of different depths and widths are executed with the previously presented \afs.
A set 4 seeds of He-Normal \cite{heDelvingDeepRectifiers2015} and 2 seeds of Glorot-Uniform \cite{glorotUnderstandingDifficultyTraining2010} weight initialisers have been generated and the results are presented in the form of mean and standard deviation.
Where applicable, experiments with L2 regularisation are also reported.

\subsubsection{Hyperbolic Tangent Function}
A \oned equation given by $f(x) = 0.5 + 0.5\tanh(5x),~x \in [-1,1]$ is used to generate two noiseless sets of equally spaced $14$- and a more challenging $7$-point cases.
The models are trained with $(n \times m) = (7 \times 120)$ and $(8 \times 240)$, for 15000 epochs with batch size 3.
The learning rate is scheduled in the range $[1e-3, 1e-6]$ with factor 0.5, patience 500, and cooldown 100.

\begin{table}[h!] \caption{Mean and standard deviation of the losses of different model setups for the hyperbolic tangent with $14$ training points.} \label{tanh-p14_table}
\centering
\begin{tabular}{l
    >{\centering\arraybackslash}p{1.5cm}
    >{\centering\arraybackslash}p{1.5cm}
    >{\centering\arraybackslash}p{1.5cm}
    >{\centering\arraybackslash}p{1.5cm}
}
\toprule
\textbf{Activation} 		& \multicolumn{2}{c}{\textbf{Training MAE} $\times 10^{-6}$} & \multicolumn{2}{c}{\textbf{Diffusion MSE} $\times 10^{-2}$} \\
\cmidrule(lr){2-3} \cmidrule(lr){4-5}
									& $n=7$		& $n=8$		& $n=7$		& $n=8$ \\
\midrule
ELU                       & (15;7)    & (18;8)   & (39;14)   & (80;98)   \\
Leaky ReLU $(\alpha=0.2)$ & (10;3)    & (15;5)   & (126;56)  & (156;59)  \\
LELU $(\beta=0.3)$        & (19;12)   & (27;33)  & (16;4)    & (25;4)    \\
LELU $(\beta=0.4)$        & (19;10)   & (30;99)  & (18;3)    & (22;6)    \\
LELU $(\beta=0.6)$        & (22;34)   & (63;57)  & (14;3)    & (18;5)    \\
ReLU                      & (7;3)     & (7;3)    & (216;59)  & (275;34)  \\
ReLU-L2                   & (5;15)    & (19;23)  & (107;3)   & (108;7)   \\
SiLU                      & (10;3)    & (12;8)   & (17;0)    & (17;0)    \\
SiLU-L2                   & (171;118) & (136;89) & (17;2)    & (17;2)    \\
Softplus                  & (43;14)   & (65;15)  & (18;0)    & (36;14)   \\
Tanh                      & (37;39)   & (69;28)  & (249;126) & (272;528) \\
\bottomrule
\end{tabular}
\end{table}

\begin{table}[h!] \caption{Mean and standard deviation of the losses of different model setups for the hyperbolic tangent with $7$ training points.} \label{tanh-p7_table}
\centering
\begin{tabular}{l
    >{\centering\arraybackslash}p{1.5cm}
    >{\centering\arraybackslash}p{1.5cm}
    >{\centering\arraybackslash}p{1.5cm}
    >{\centering\arraybackslash}p{1.5cm}
}
\toprule
\textbf{Activation} 		& \multicolumn{2}{c}{\textbf{Training MAE} $\times 10^{-6}$} & \multicolumn{2}{c}{\textbf{Diffusion MSE} $\times 10^{-2}$} \\
\cmidrule(lr){2-3} \cmidrule(lr){4-5}
									& $n=7$		& $n=8$		& $n=7$		& $n=8$ \\
\midrule
ELU                       & (27;12) & (22;32)   & (41;8)  & (54;50)  \\
Leaky ReLU $(\alpha=0.2)$ & (7;7)   & (10;6)    & (37;38) & (54;39)  \\
LELU $(\beta=0.3)$        & (17;14) & (57;54)   & (14;4)  & (28;10)  \\
LELU $(\beta=0.4)$        & (40;15) & (60;36)   & (8;4)   & (9;2)    \\
LELU $(\beta=0.6)$        & (35;24) & (64;33)   & (14;6)  & (16;5)   \\
ReLU                      & (6;5)   & (6;4)     & (37;63) & (53;115) \\
ReLU-L2                   & (8;5)   & (2;2)     & (72;4)  & (74;4)   \\
SiLU                      & (9;5)   & (10;5)    & (5;4)   & (7;3)    \\
SiLU-L2                   & (5;106) & (170;142) & (9;1)   & (8;0)    \\
Softplus                  & (21;11) & (51;28)   & (58;35) & (71;35)  \\
Tanh                      & (13;8)  & (10;22)   & (96;42) & (86;39) \\
\bottomrule
\end{tabular}
\end{table}

\begin{figure}[ht!]
  \centering
  \subfloat[7 training points.]{
      \includegraphics[width=0.625\textwidth]{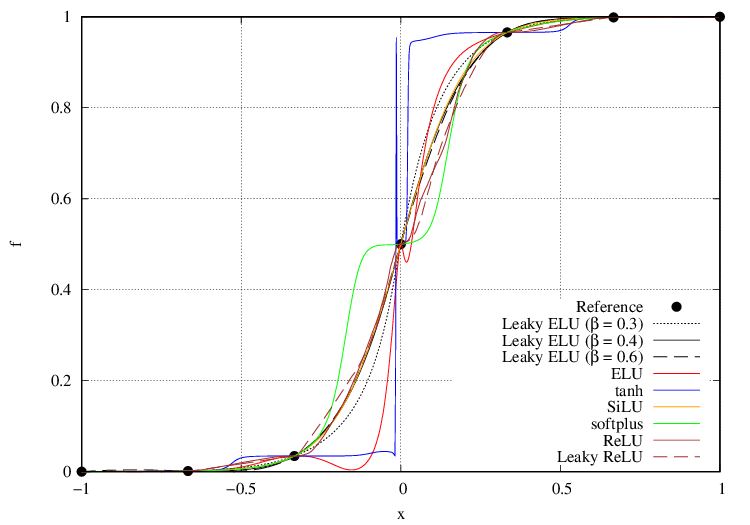}\label{tanh_funcs_a}}
  \vfill
  \subfloat[14 training points.]{
      \includegraphics[width=0.625\textwidth]{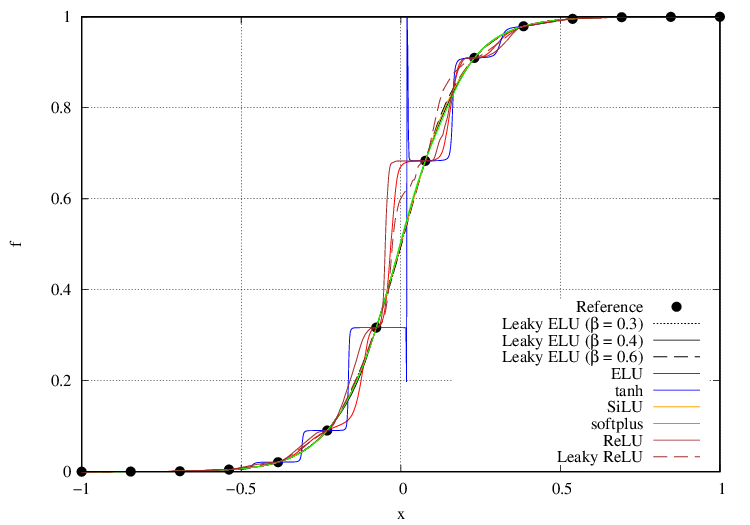}}
  \caption{Trained hyperbolic tangent function plots for the different activation functions for the He-Normal initialisation (no seed).}
  \label{tanh_funcs}
\end{figure}

Tables \ref{tanh-p14_table} and \ref{tanh-p7_table} present the training and diffusion\footnote{In this case for which the function values approach zero, the denominators in Eqs.\ \ref{truescore1d} and \ref{staggeredlaplacian} are not implemented.} losses for $14$ and $7$ training points, respectively.
As expected, the models tend to perform better with the larger number of training points, however the errors are not necessarily much smaller.
The models on the lower end of MAEs are generally tending to memorise the training points, as seen in the function plots in Fig.\ \ref{tanh_funcs}, so the performance of the models is fundamentally gauged by the diffusion metric.

SiLU and Softplus present adequate diffusion metrics for this hyperbolic tangent function, closely followed by LELU.
Softplus however loses performance with the deeper model and with the lower number of points.
All other \afs perform rather poorly, and the application of regularisation (strength $1e-4$) does not improve or even worsens the performance.
For this particular problem, LELU can perform closely to SiLU at the expense of higher training MAE.
It is expected that the optimal parameter $\beta$ of LELU is problem \textit{and} model dependent, which can be observed in the presented results, and that is an intended property.

\subsubsection{Exponential Function}
A \oned exponential equation given by $f(x) = 0.1^x,~x \in [-1,1]$ is used to generate a set of $12$ equally spaced, noiseless training data points.
The models are trained with $(n \times m) = (7 \times 240)$ and $(8 \times 240)$, for 15000 epochs with batch size 3.
The learning rate is scheduled in the range $[1e-3, 2e-7]$ with factor 0.5, patience 500, and cooldown 100.

\begin{table}[h!] \caption{Mean and standard deviation of the losses of different model setups for the exponential function with $12$ training points.} \label{expot_table}
\centering
\begin{tabular}{l
    >{\centering\arraybackslash}p{1.5cm}
    >{\centering\arraybackslash}p{1.5cm}
    >{\centering\arraybackslash}p{1.5cm}
    >{\centering\arraybackslash}p{1.5cm}
}
\toprule
\textbf{Activation} 		& \multicolumn{2}{c}{\textbf{Training MAE} $\times 10^{-6}$} & \multicolumn{2}{c}{\textbf{Diffusion MSE} $\times 10^{-3}$} \\
\cmidrule(lr){2-3} \cmidrule(lr){4-5}
							& $n=7$	& $n=8$	& $n=7$	& $n=8$ \\
\midrule
ELU                       & (21;62)   & (20;15)   & (12;9) & (19;16) \\
Leaky ReLU $(\alpha=0.2)$ & (10;9)    & (11;14)   & (14;4) & (15;3)  \\
LELU $(\beta=0.3)$        & (20;6)    & (36;25)   & (2;1)  & (3;2)   \\
LELU $(\beta=0.4)$        & (44;57)   & (40;99)   & (2;1)  & (2;1)   \\
LELU $(\beta=0.6)$        & (38;227)  & (52;184)  & (2;0)  & (1;0)   \\
ReLU                      & (8;6)     & (7;8)     & (21;4) & (18;4)  \\
ReLU-L2                   & (6;1)     & (7;2)     & (18;8) & (24;7)  \\
SiLU                      & (17;95)   & (12;75)   & (15;8) & (13;7)  \\
SiLU-L2                   & (155;683) & (104;567) & (8;46) & (7;5)   \\
Softplus                  & (115;111) & (97;166)  & (2;1)  & (4;0)   \\
Tanh                      & (11;5)    & (157;165) & (33;9) & (24;23) \\
\bottomrule
\end{tabular}
\end{table}

\begin{figure}[ht!]
  \centering
  \subfloat[Depth 7.]{
      \includegraphics[width=0.625\textwidth]{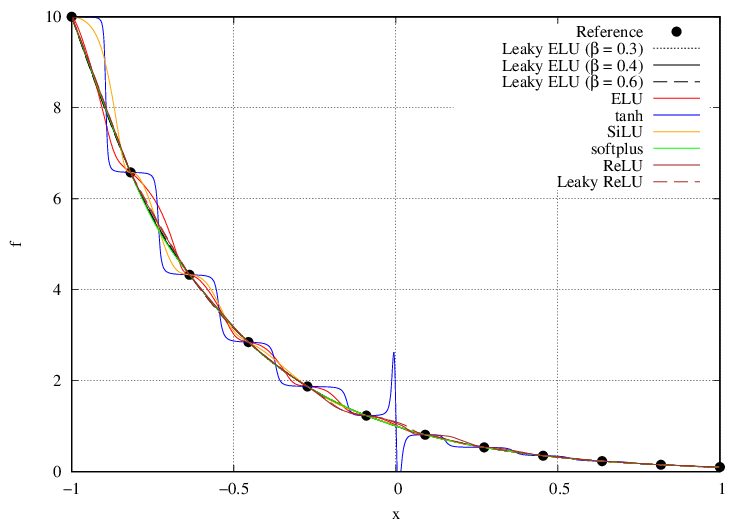}}
  \vfill
  \subfloat[Depth 8.]{
      \includegraphics[width=0.625\textwidth]{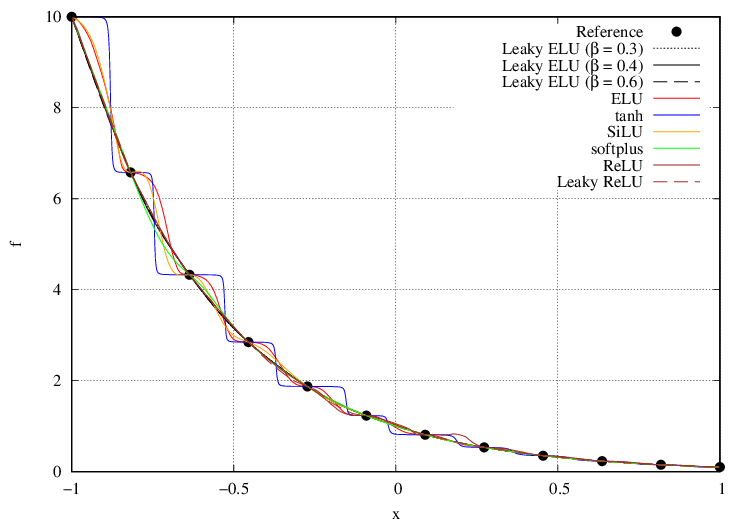}}
  \caption{Trained exponential function plots for the different activation functions.}
  \label{expot_funcs}
\end{figure}

Table \ref{expot_table} presents the training and diffusion losses of the trained models.
As in the previous case, the lower the MAE, the higher is the tendency of the model to memorise the training data with a resulting increase in the diffusion metric (see Fig.\ \ref{expot_funcs}).
Different levels of overfitting are observed with the classical \afs (including regularisation) in this setup, whereas generally robust results are obtained with LELU.
Contrary to the previous observations with the hyperbolic tangent results, both SiLU and Softplus show a poorer performance in terms of training MAE or diffusion MSE.
The LELUs are able to adequately meet both low MAE and diffusion errors, with the tendency for slightly better overall performance with lower $\beta$ due to lower MAE.

\subsubsection{Exponential Function and a Single Neuron}
\begin{figure}[th]
  \centering
  \includegraphics[width=0.65\textwidth]{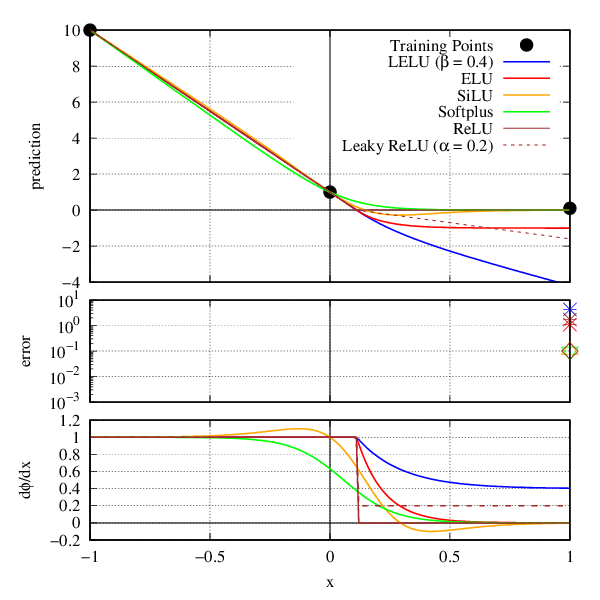}
  \caption{Predictions with a single neuron trained for the three data points.}
  \label{singleneuron}
\end{figure}

The previous problem is reduced to a set of three training points, $x_i = (-1, 0, 1)$, which a single neuron shall learn.
This simple exercise can be performed by hand so two of the three points can be exactly matched.
The obtained neuron parameters are used to infer a function as shown in Fig.\ \ref{singleneuron}, plotted along the \af derivatives and training errors.

In terms of training error (at the third point), ReLU, SiLU and Softplus produce smaller errors whereas the LELU has the largest.
However, ReLU, ELU, SiLU and Softplus tend to null gradients whereas the Leaky ReLU and LELU produce a strong signal that indicates that that behaviour cannot be coded -- the task therefore requires and should be taken by the next of kin on a larger model.

From the perspective of learning nonlinear regression tasks, such weak signal is less then ideal.
The gradient and Hessian of the loss can be computed by hand for this simple problem (motivated by \cite{constantinescuApproximationInterpolationDeep2024}).
This operation shows that the weak \afs produce a vanishing loss gradient and higher conditioned Hessians, which is indicative of the tendency to poor learning and overfitting.
The LELU on the other hand produces a clear gradient and a lower conditioned Hessian.
With this simple exercise, the author intends to anecdotally show that the higher ``flexibility'' of the \af is incurring in a weak signal for subsequent neurons and is a potential cause of instability showing itself as overfitting.

\subsubsection{Other Findings with the Unidimensional problems}
\begin{figure}[h]
  \centering
  \subfloat[Hyperbolic tangent, with 7 points.]{
      \includegraphics[width=0.475\textwidth]{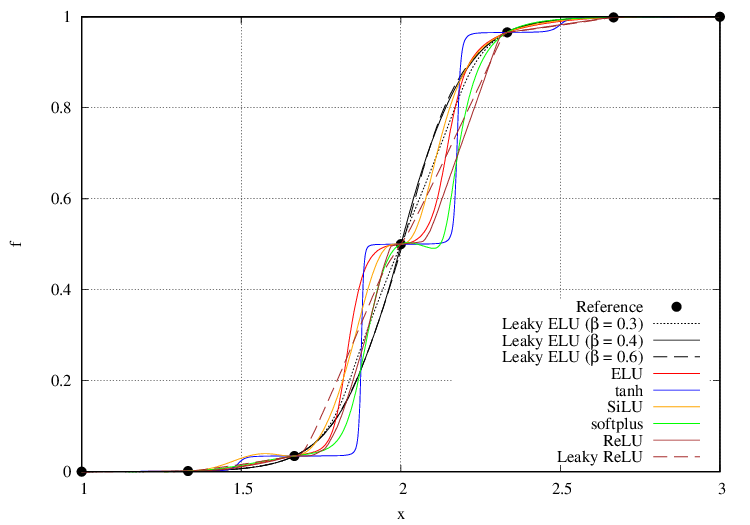}}
  \hfill
  \subfloat[Exponential function, depth 8.]{
      \includegraphics[width=0.475\textwidth]{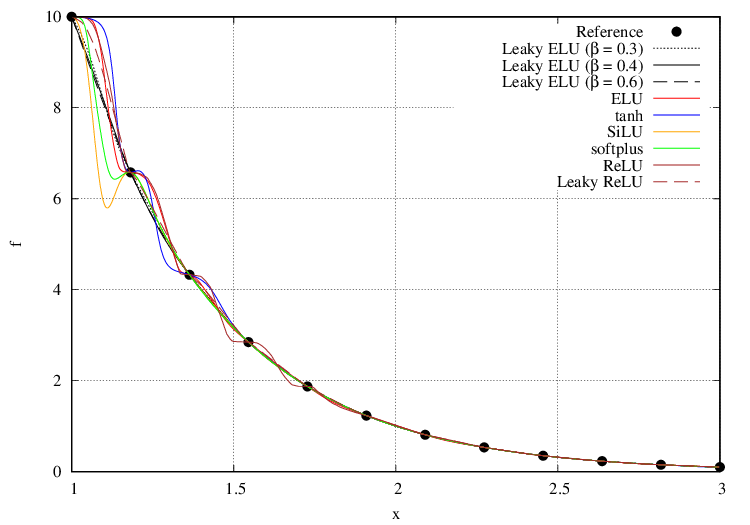}}
  \caption{Trained function plots for the different activation functions on a positive range of $x$.}
  \label{expot_funcs_shift}
\end{figure}

It is interesting to note that some \afs struggle around $x = 0$ in both case studies, even when this point is \textit{not} a training data.
A thorough explanation is challenging for a large model \cite{huMeasuringModelComplexity2020}, but the author believes this can be partially understood by a similar analysis of the gradient as in the previous subsection.
The loss-gradient is proportional to the summation of the gradient of the \af times the training point coordinate.
If the training point coordinate tends to zero, or adjacent training points cross $x=0$, this can further induce the loss gradient summation to vanish or ``flicker'', and to destabilize the learning on an already unstable model setup.
This behaviour is confirmed with a simple shift in $x$ by two units, as shown in Fig.\ \ref{expot_funcs_shift}, in which the local instability around $x=0$ is removed for the unstable models.
Note that SiLU actually lost performance and presents overfitting, opposed to the behaviour shown in Fig.\ \ref{tanh_funcs_a}, which corroborates the borderline stability of the model.

Another observation is that C-1 discontinuous \afs generate ``polyline'' trained models.
In these and some other setups not shown for conciseness, (leaky) ReLUs rather often default to piecewise linear connection between the training points or at least connect them with several segments of polyline.
Such patterns are shown in the results in Figs.\ \ref{tanh_funcs}, \ref{expot_funcs} and \ref{expot_funcs_shift}, as well for the motor model to be discussed in what follows.
Interestingly, in none of the cited ReLU-related literature in the introduction, the herein reported ``polyline'' pattern is commented.

The adopted canonical problems indicate that classical \afs lack robustness with the problem setup for nonlinear regressions tasks when trained by large models.
The proposition of the objectives in Sect.\ \ref{afs} for the LELU construction aims at mitigating the reported issues.
LELU is able to achieve adequate results without heavily incurring in such penalties for the presented canonical problems.
It can robustly learn the nonlinear regression task without memorising the training data, at the expense of slightly higher training MAEs, resembling an inherent but consistent auto-regularisation property of the \af.
This behaviour is further confirmed on the more complex model discussed in what follows.

\subsection{Electric Motor Model} \label{sheselectric}
\begin{figure}[t]
	\centering
	\includegraphics[clip, trim=3cm 3cm 3cm 3cm, width=11cm]{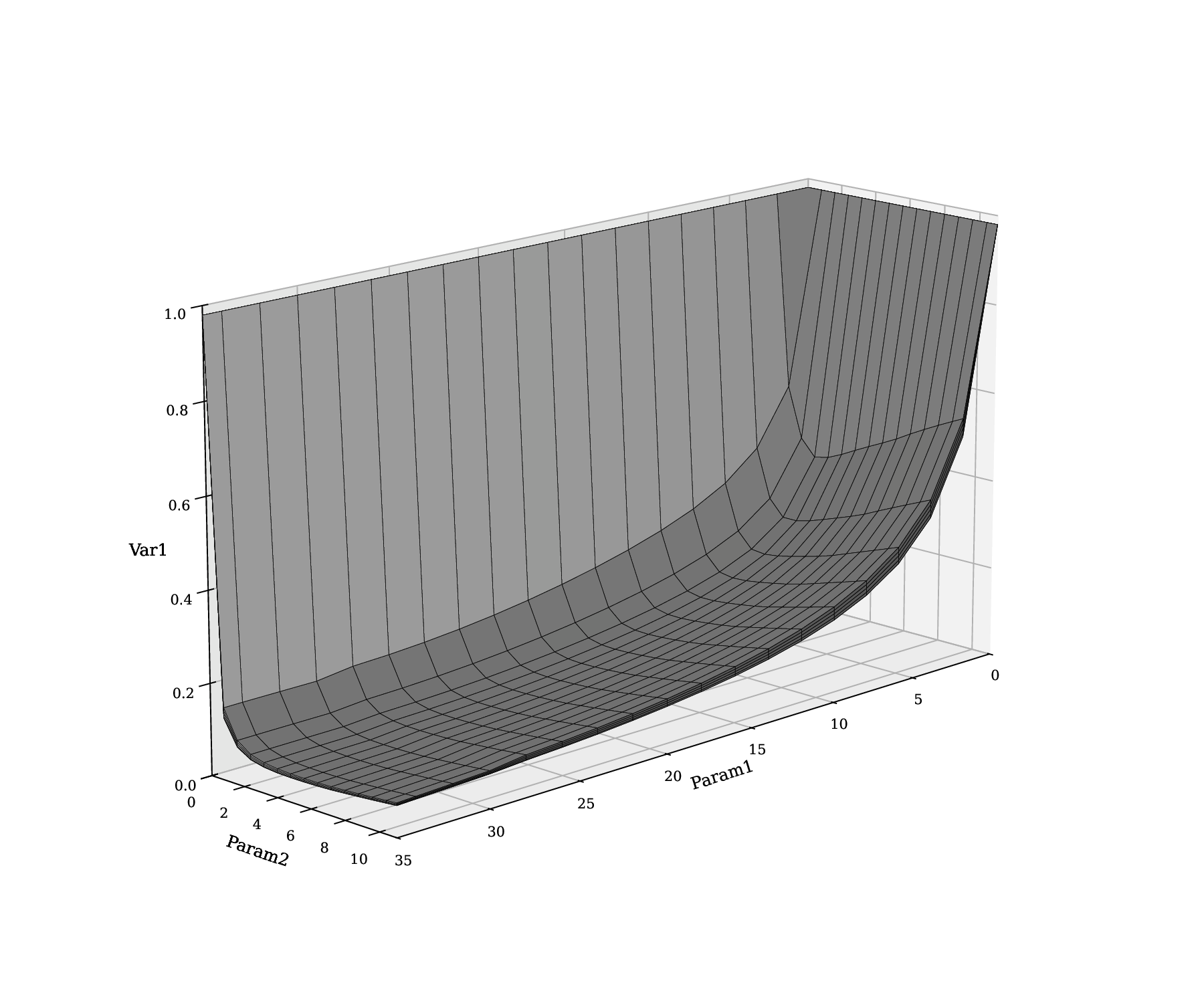}
	\caption{Wireframe view of the motor model mesh with the third dimension shown as the ``thickness'' of the wireframe. The third dimension collapses at the lower borders.}
	\label{envelope}
\end{figure}

The noise-free system model described in the introduction is composed of a \threed mesh of \textit{(param1, param2, param3)} = $(19 \times 15 \times 5)$ points, shown in Fig.\ \ref{envelope}.
The problem shows a fast change of behaviour along $param1$ and an even faster one with $param2$, which combine at one corner of the operational envelope of the motor.
The behaviour with $param3$ shows as the ``thickness'' of the wireframe, and it changes with both other parameters and collapses at their lower borders.
A nonlinear feature can also be observed approaching the maximum limits of the $param1$ range.

The dataset is normalised so as to keep unitary mesh spacing in all dimensions to avoid cell skewness (which also favours the computation of the diffusion differentials), and kept in the positive range to support the more sensitive \afs.
For the current study the trained data must exactly reproduce the noise-free training information \cite{bigarellaPreventionOverfittingMeshStructured2025}.
A noiseless training is generally more challenging and is an additional reason for this choice in the current exercise.

A numerical experiment with the implemented\footnote{Source files available in \url{http://github.com/ebiga/gpr_aniso_trials}} NN with combinations of depth $n = [6, 7, 8]$ and width $m = [80, 120, 240, 360]$ (fixed per layer) is executed.
The model is trained for 9000 epochs with batch size 35.
The learning rate is scheduled in the range $[1e-3, 1e-5]$ ($5e-4$ for the models with width $360$ and the $8 \times 240$ case) with factor 0.5, patience 250, and cooldown 50.
ELU, SiLU, and LELU ($\beta=0.4$) are run with 4 seeds of He-Normal and 1 seed of Glorot-Uniform weight initialisers for the $7 \times 120$ and $8 \times 120$ cases, and the standard deviations are presented as error bars.
Where applicable, experiments with L2 regularisation are also reported.

\begin{figure}[ht]
  \centering
  \subfloat[MAE.]{
      \includegraphics[clip, trim=0cm 0cm 4cm 0cm, height=5.5cm]{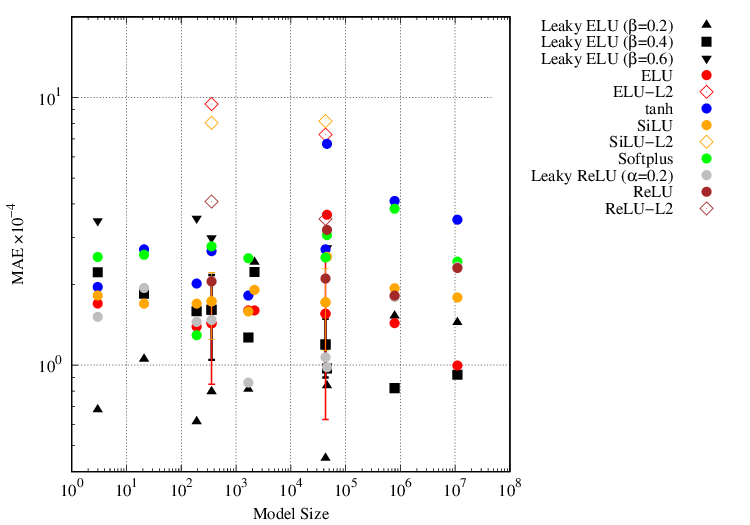}}
  \hfill
  \subfloat[Diffusion SE.]{
      \includegraphics[clip, height=5.5cm]{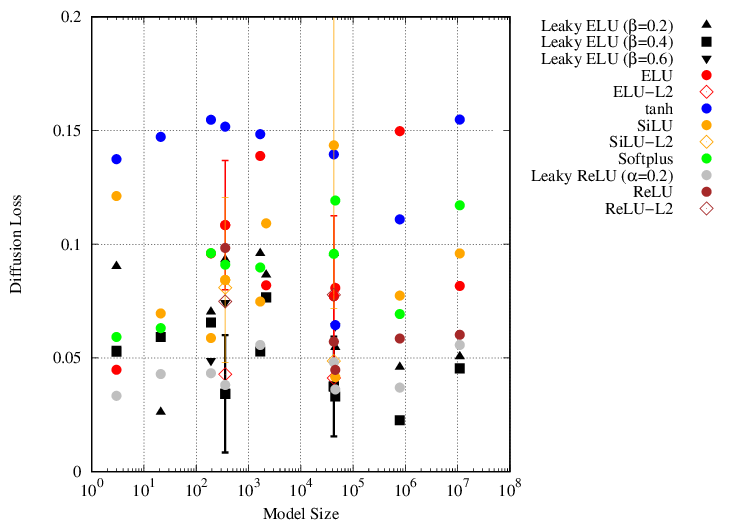}}
  \caption{Electric motor model losses for different \afs and model sizes indexed by $m^n/1e12$.}
  \label{motor_scatter_losses}
\end{figure}
Figure \ref{motor_scatter_losses} shows the MAE and diffusion metrics for the different model sizes and \afs.
The tendency to lose generalisation shows as a lack of ``convergence'' of the losses with the increase in model size.
For this complex problem the mean errors do not show large differences between the ``good'' and ``bad,'' so visual inspection guided by the results of the diffusion loss is necessary.

\begin{figure}[th!]
  \centering
  \subfloat[LELU $(\beta=0.2)$]{\includegraphics[width=0.475\textwidth]{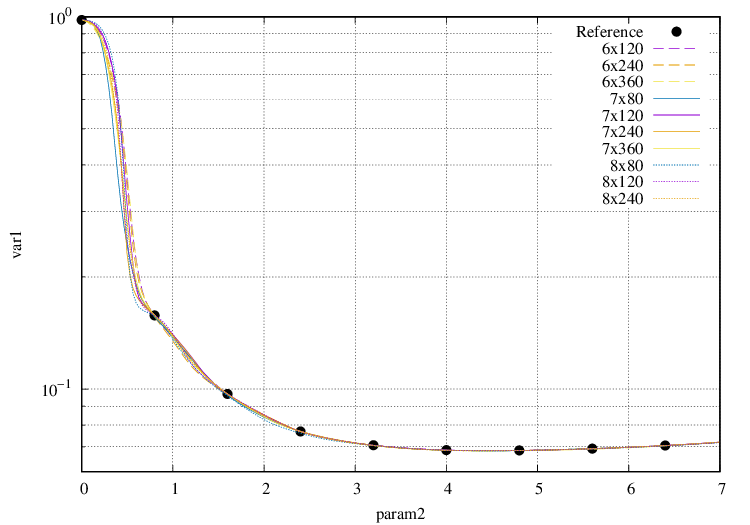}}
  \hfill
  \subfloat[LELU $(\beta=0.4)$]{\includegraphics[width=0.475\textwidth]{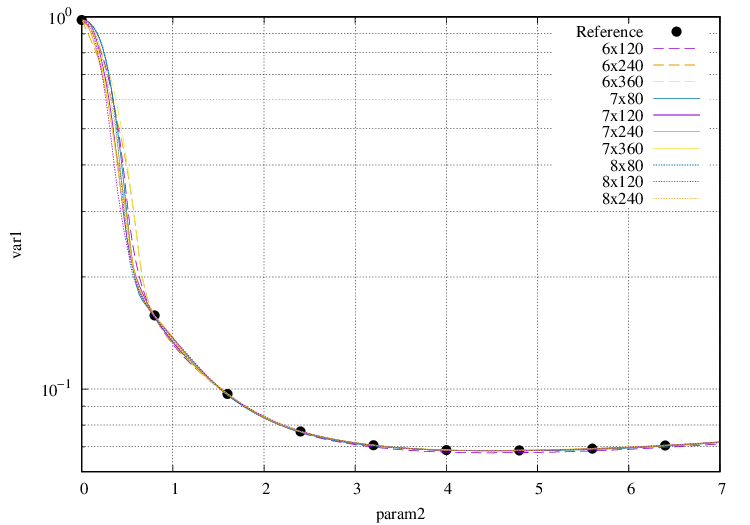}}
  \vfill
  \subfloat[ELU]{\includegraphics[width=0.475\textwidth]{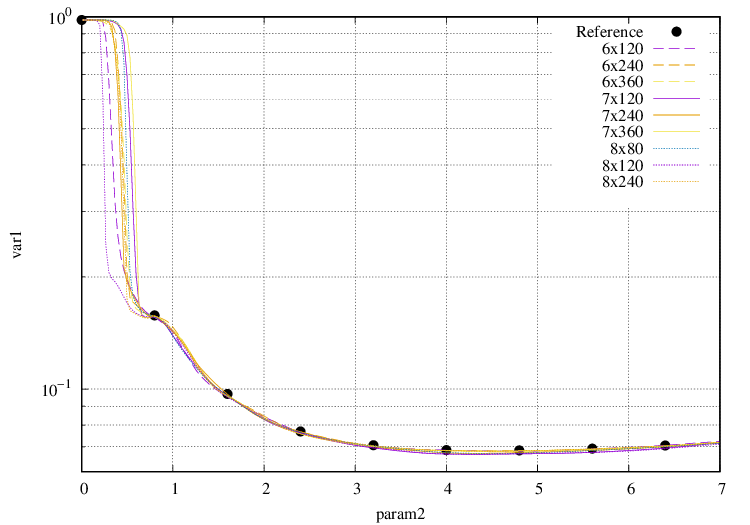}}
  \hfill
  \subfloat[SiLU]{\includegraphics[width=0.475\textwidth]{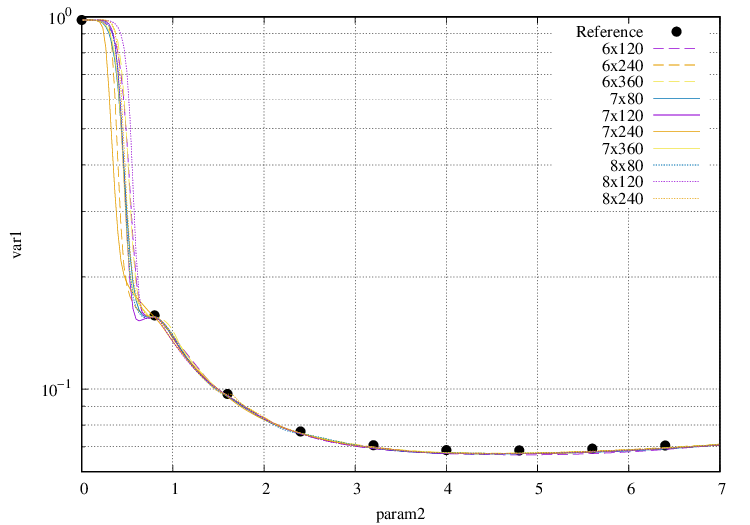}}
  \vfill
  \subfloat[Softplus]{\includegraphics[width=0.475\textwidth]{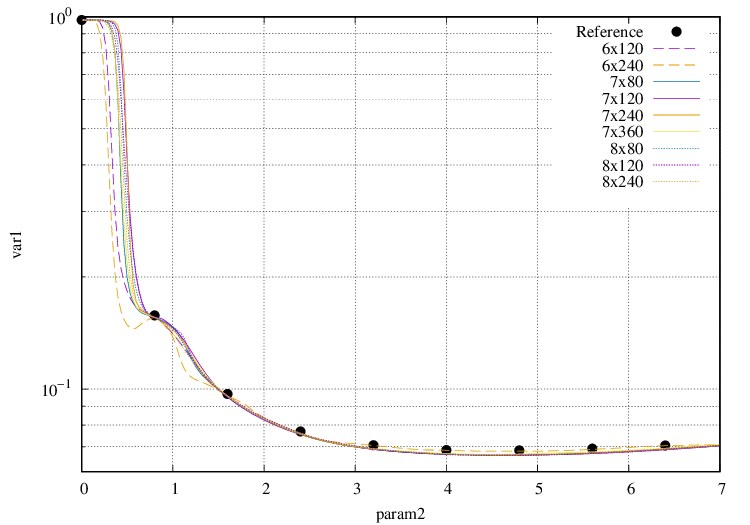}}
  \hfill
  \subfloat[Leaky ReLU $(\alpha=0.2)$]{\includegraphics[width=0.475\textwidth]{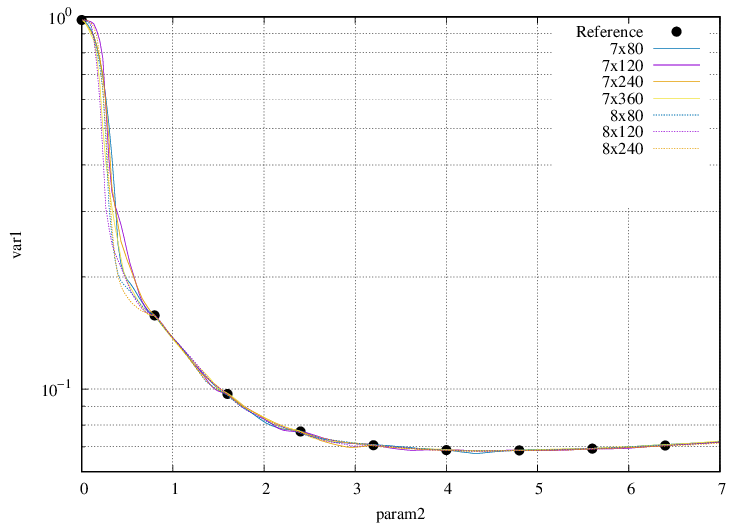}}
  \caption{Trained electric motor model predictions sliced at $(param1, param3) = (28, 0.8)$ for different activation functions and model sizes.}
  \label{motor_param2}
\end{figure}

\begin{figure}[th!]
  \centering
  \subfloat[LELU $(\beta=0.2)$]{\includegraphics[width=0.475\textwidth]{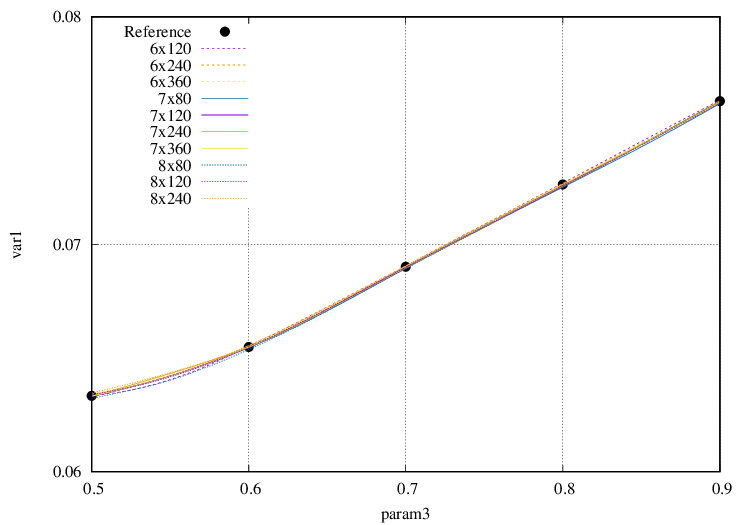}}
  \hfill
  \subfloat[LELU $(\beta=0.4)$]{\includegraphics[width=0.475\textwidth]{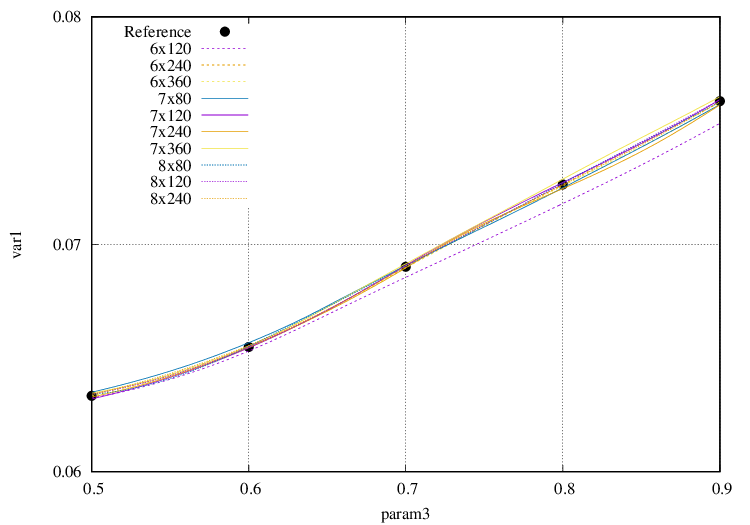}}
  \vfill
  \subfloat[ELU]{\includegraphics[width=0.475\textwidth]{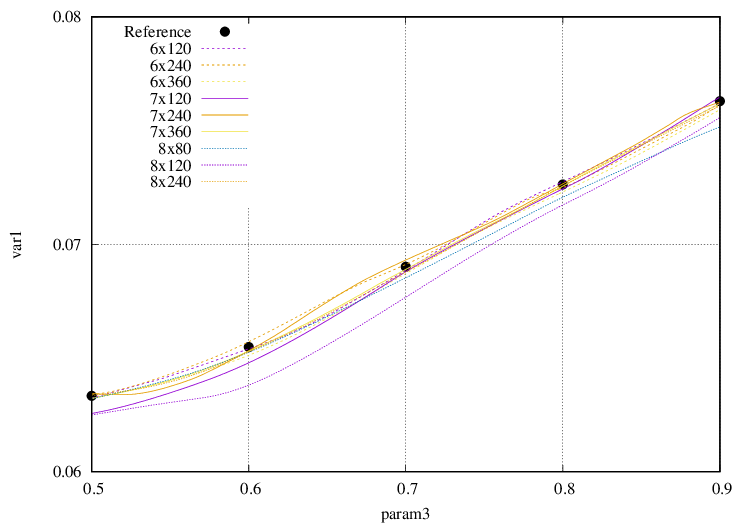}}
  \hfill
  \subfloat[SiLU]{\includegraphics[width=0.475\textwidth]{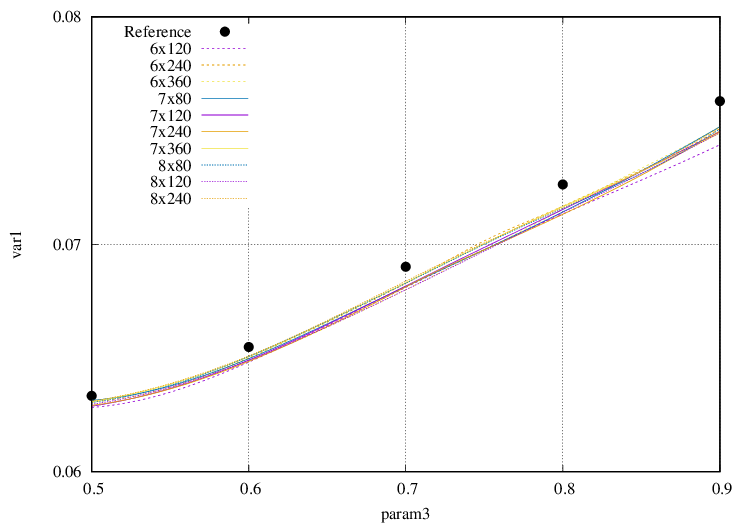}}
  \vfill
  \subfloat[Softplus]{\includegraphics[width=0.475\textwidth]{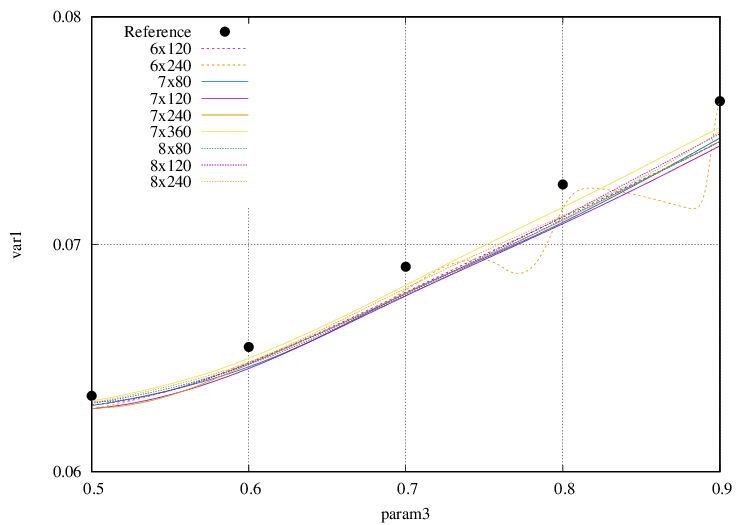}}
  \hfill
  \subfloat[Leaky ReLU $(\alpha=0.2)$]{\includegraphics[width=0.475\textwidth]{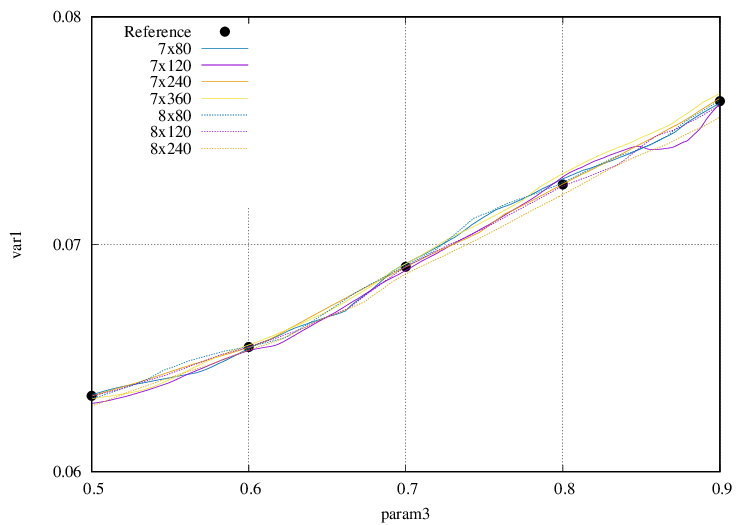}}
  \caption{Trained electric motor model predictions sliced at $(param1, param2) = (28, 7.2)$ for different activation functions and model sizes.}
  \label{motor_param3}
\end{figure}

In overall, similar behaviours as with the canonical problems are also observed in this case.
In special, the most promising classical \afs SiLU and Softplus also present poor performance, consistent with the slice plots in Figs.\ \ref{motor_param2} and \ref{motor_param3}.
LELU successfully achieves lower losses amongst the tested \afs and is also more robust to model size.
As expected, the losses with different values of $\beta$ can be sensitive to model parameters.
As an example, $\beta = 0.2$ presents better results for smaller models and loses performance with larger ones, whereas $\beta = 0.6$ presents the opposite trend.
LELU with $\beta = 0.4$ presents almost iso-performance with model parameters, and the sensitivity (error bars) to initialisation is also the smallest.
In all cases, the results are more promising than any other \af.

Figure \ref{motor_param2} shows slices along $param2$, which is the fast-changing variable.
The tanh and ReLU results are not shown due to poor performance.
The region $param2 = [0,2]$, where the highest slopes are seen, presents a tough task for the \afs.
Considerable ``localised'' overfitting is observed for the ELU, SiLU and Softplus.
Interestingly, the less flexible \textit{leaky} \afs show lower overfitting content and also less dependency on the model size.
This nice behaviour prompted the current effort into assembling this property into a new \af.
The Leaky ReLU results are ``kinky'' though and such behaviour is not ideal.
The proposed LELU with $\beta = 0.4$ shows a robust performance, indicating that the lower flexibility and smoothness are indeed advantageous.

Figure \ref{motor_param3} shows similar slices along the tight- and slow-changing $param3$.
This slice allows for interesting behaviours to be more closely assessed, \ie, the \textit{spread} of the \af with different model sizes and the \textit{capacity to code} small values.
The less flexible \textit{leaky} \afs clearly show tighter and more representative capacity.
SiLU and Softplus are also able to demonstrate a tighter spread, albeit with an offset error to the training data, whereas ELU is generally poorly behaved.

By inspection of \af graphs in Fig.\ \ref{activ_fks}, it can be seen that both SiLU and Softplus present a wider range of slowly changing gradient, or in other words, ELU has a sharper gradient change.
In the SiLU case, the gradient is actually never zero in the range plotted in Fig.\ \ref{activ_fks}.
This observation has further prompted the effort into coding this property into the LELU.
As seen in Fig.\ \ref{activ_fks}, this \af slowly changes slopes and is therefore smoother than other \afs, thereby enabling a stronger a priori fitness for generalisable nonlinear regressions.

\begin{figure}[th!]
  \centering
  \subfloat[$6\times120$]{
  	\includegraphics[width=0.475\textwidth]{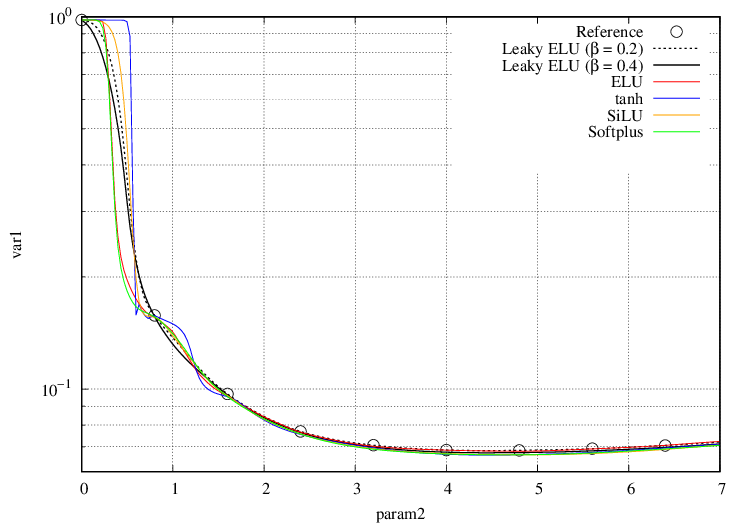}
  	\includegraphics[width=0.475\textwidth]{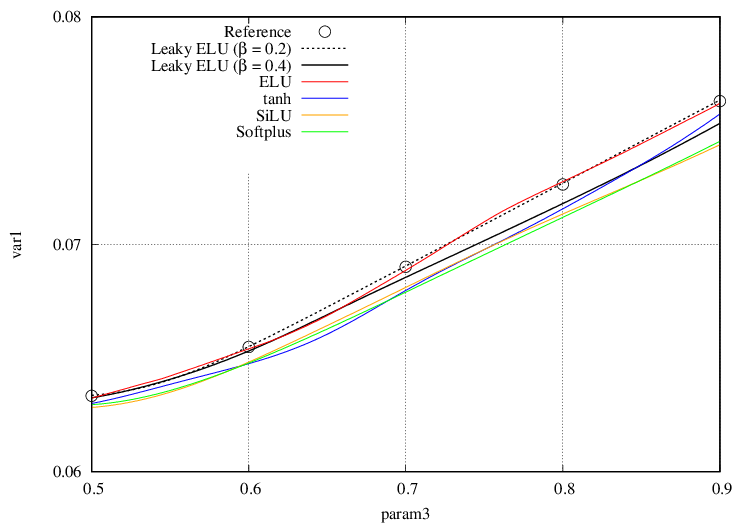}}
  \vfill
  \subfloat[$7\times120$]{
  	\includegraphics[width=0.475\textwidth]{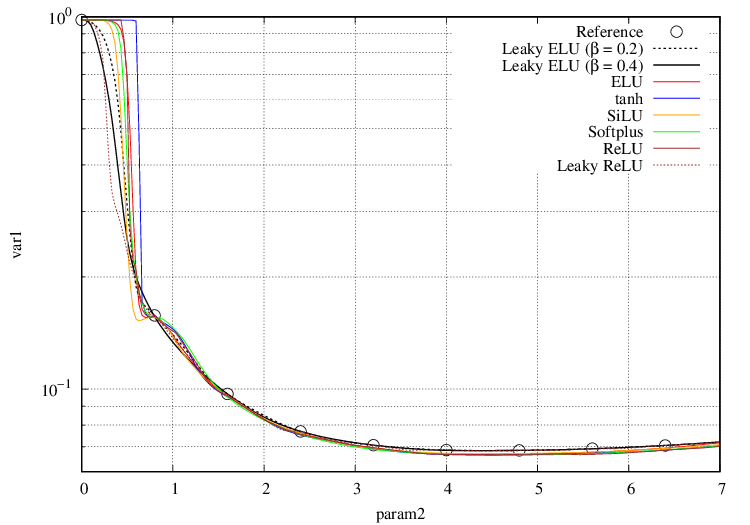}
  	\includegraphics[width=0.475\textwidth]{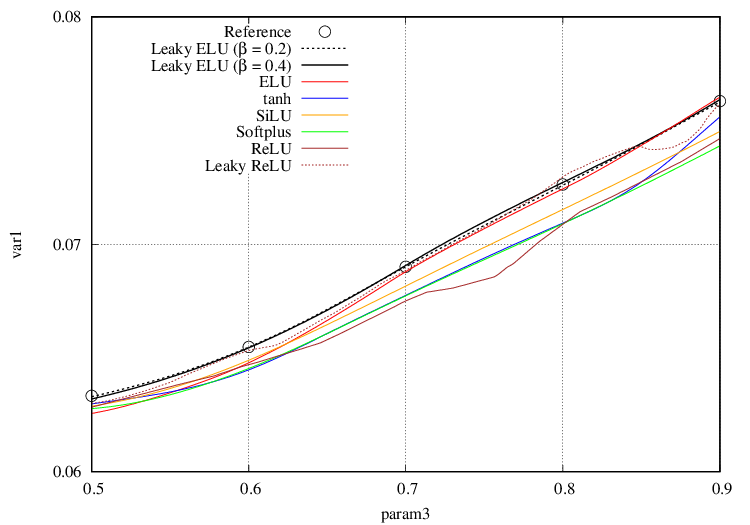}}
  \vfill
  \subfloat[$8\times120$]{
  	\includegraphics[width=0.475\textwidth]{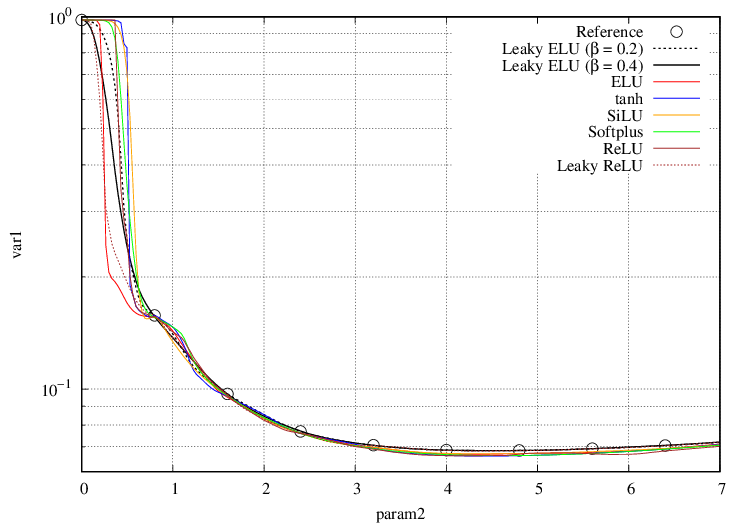}
  	\includegraphics[width=0.475\textwidth]{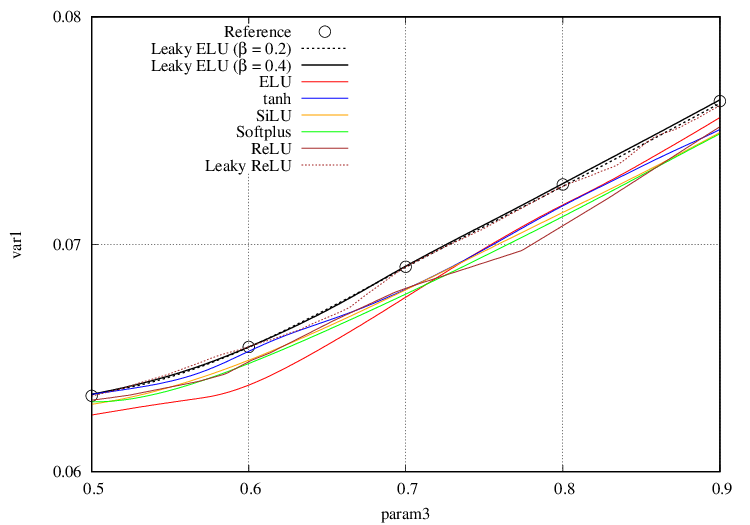}}
  \caption{Model size effect on the electric motor predictions for different activation functions.}
  \label{motor_models_lelu04}
\end{figure}
Upon careful visual inspection of each NN result (see Fig.\ \ref{motor_models_lelu04}), depth 6 is not enough to capture all features, whereas depth 7 is able to do so but the MAE still indicates some distance (error) to the training points.
Generally at least width 120 is necessary for acceptable results.

\begin{figure}[th!]
  \centering
  \subfloat[]{
  	\includegraphics[width=0.475\textwidth]{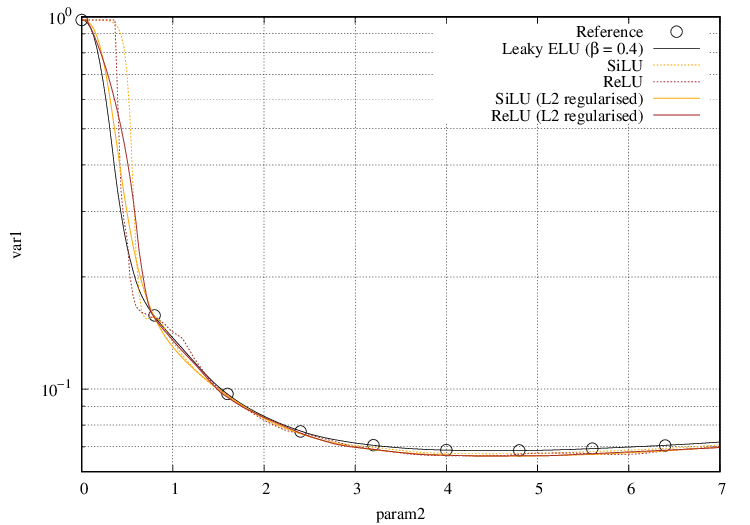}}
  \hfill
    \subfloat[]{
  	\includegraphics[width=0.475\textwidth]{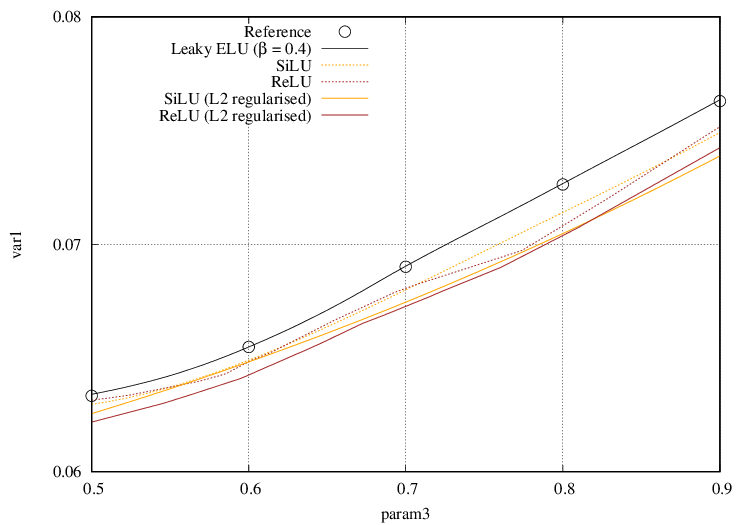}}
  \caption{Regularisation effect on the electric motor predictions for the $8\times120$ model.}
  \label{motor_models_reg}
\end{figure}
Figure \ref{motor_models_reg} shows slices with L2 (strength $1e-4$) regularisation to reduce the overfitting tendency of ReLU and SiLU.
As corroborated by the results in Fig.\ \ref{motor_scatter_losses}, the diffusion loss indicates a reduction in overfitting at the expense of an increase in the training MAE by more than a factor of 6.
The performance does not improve over the LELU ($\beta = 0.4$) and ReLU still keeps its ``kinky'' piecewise-linear predictions.

Another strategy is to implement a change of variables, for instance a power correction $\tilde y = y^{0.1}$, to reduce the large differences in the max and min values.
A model trained on this function learns faster and easier, however the resulting errors when brought back to the original scale are also very large, which defeats the purpose.
As already observed with the canonical problems, the LELU seems to provide adequate results without those drawbacks and extra steps.

\section{Concluding Remarks}
A parametric activation function for nonlinear regressions tasks that is monotonically smooth, with a non-null gradient at the negative region, is developed.
The proposed \af is termed Leaky Exponential Linear Unit (or Leaky ELU, or LELU).
The non-null gradient is proposed as a means to control an \adhoc score of the flexibility of an \af and is a trainable parameter.
Available nonlinear \afs have generally high flexibility scores which seem to be associated with proneness to overfitting.
A null flexibility score can only represent linear data, on the other hand.

Smoothness is sought to avoid coding discontinuities at the predictions of the trained model.
This is shown to be a result of the ReLU and Leaky ReLU \afs.
The Leaky ReLU however has a trainable flexibility and this is shown to support the claim of beneficial and robust behaviour of this property for nonlinear regression tasks.

The results with the LELU are promising for highly nonlinear regressions tasks and large models.
Its properties implicitly produce a behaviour similar to regularisation without strong penalisation of the training errors.
The LELU does not show evidence of trading off training error with the overfitting/testing metric, at least not as strongly as other \afs.
It also shows less sensitivity with model size and training sample size for nonlinear regression tasks.
The parameter $\beta$ is proposed so it can be dedicatedly trained for better generalisation for different problems.

Another proposition in the current work is a metric to identify overfitting.
The method is based on a modified Laplace operator and tests for oscillation in the interior of the training mesh cells.
A true label is provided by the modified Laplace operator applied to the original training nodes, which acts as a measure of the ``entropy-in-features'' contained in the training data.
A \textit{trained} model must limit (ideally avoid) the amount of extra features it determines based on its diffusion metric.
The testing points must be created on a staggered fashion to be able to gauge any excess entropy in the interior of the mesh cells.
With the proposed method, the model can be trained on all available training points.

The author expects the proposed methods to be generalisable to other machine learning or dataset strategies.
The observed properties of the proposed LELU should reduce the workload to select a good model setup and is as such a highly desirable behaviour.
The Laplace operator with its measure of the diffusion of a system provides a powerful tool for detection (and prevention) of overfitting in data regression. 
They should provide a sure hand in the process of implementing a model by bringing more robustness and clearer metrics.

\bibliographystyle{elsarticle-num}
\bibliography{biblio}

\end{document}